%% file: SamiskOCR.tex
\documentclass[11pt]{article}
\usepackage{nodalida2025}
\usepackage[T1]{fontenc}
\usepackage{times}
\usepackage{url}
\usepackage{latexsym}
\usepackage{hyperref}
\usepackage{multirow}
\usepackage[table,xcdraw]{xcolor}
\usepackage{colortbl} 
\usepackage{booktabs}
\usepackage{makecell}
\usepackage{amsmath}
\usepackage{amssymb}
\usepackage[utf8]{inputenc}
\usepackage{threeparttable}
\usepackage{stmaryrd}
\usepackage{tipa}

\usepackage{titlesec}
\titleformat{\subsubsection}{\bfseries}{}{1em}{}

\aclfinalcopy % Uncomment this line for the final submission

\makeatletter
\newif\if@anonymize
\@anonymizefalse  % Uncomment to show text

\newcommand{\anonymize}[2]{%
  \if@anonymize
    #1  % Hide the text
  \else
    #2  % Show the anonymized alias
  \fi
}
\makeatother

\usepackage{listings}
\usepackage{todonotes}
\usepackage{threeparttable}
\usepackage{subcaption}
\usepackage{longtable}
\usepackage[capitalize]{cleveref}
\usepackage{caption}
\usepackage{array}
\DeclareCaptionFormat{customlongtable}{#1\\#3}
\newcommand{\emphpar}[1]{\paragraph{\emph{#1}}}

\title{Comparative analysis of optical character recognition methods for Sámi texts from the National Library of Norway}

\author{Tita Enstad$^1$, Trond Trosterud$^2$, Marie Iversdatter Røsok$^1$, Yngvil Beyer$^1$, Marie Roald$^1$\\ \\
    $^1$National Library of Norway \\
    $^2$The Arctic University of Norway \\ \\
    {\tt tita.enstad@nb.no}    
    }

\date{}

\begin{document}
\maketitle

\begin{abstract}
Optical Character Recognition (OCR) is crucial to the National Library of Norway’s (NLN) digitisation process as it converts scanned documents into machine-readable text. However, for the Sámi documents in NLN's collection, the OCR accuracy is insufficient. Given that OCR quality affects downstream processes, evaluating and improving OCR for text written in Sámi languages is necessary to make these resources accessible. To address this need, this work fine-tunes and evaluates three established OCR approaches, Transkribus, Tesseract and TrOCR, for transcribing Sámi texts from NLN's collection. Our results show that Transkribus and TrOCR outperform Tesseract on this task, while Tesseract achieves superior performance on an out-of-domain dataset. Furthermore, we show that fine-tuning pre-trained models and supplementing manual annotations with machine annotations and synthetic text images can yield accurate OCR for Sámi languages, even with a moderate amount of manually annotated data.

\end{abstract}

\section{Introduction}
% das: removed reference to PostScript

Optical Character Recognition (OCR) converts scanned documents into machine-readable text, which is crucial for making digitised materials available for search and analysis. For the National Library of Norway (NLN), the OCR output, among others, facilitates search for the online library (\emph{Nettbiblioteket}\footnote{\url{https://www.nb.no/search}}) and underpins analysis tools like the DH-Lab toolbox \cite{birkenesNBDHLABCorpus2023}. However, while OCR quality is high for most Norwegian documents, it falls short for Sámi documents. The resulting text is insufficient for both search and for use in research or as a basis for language technology. 

NLN has material in five Sámi languages: North Sámi, South Sámi, Lule Sámi, Inari Sámi and Skolt Sámi. Thus, developing an accurate OCR model for Sámi texts is important for NLN's mission to store and disseminate the materials in the library collection. Furthermore, for languages with limited resources, like Sámi languages, it is vital that the available resources are accessible to be searched and used for research. This paper describes a twofold contribution towards this goal: 

\begin{enumerate}
    \item Developing an OCR model for Sámi languages that improves the transcription accuracy of Sámi text in NLN's collection.
    \item Comparing different OCR approaches in terms of transcribing smaller languages such as languages in the Sámi family. 
\end{enumerate}

\section{Background}

\subsection{Sámi languages in the National Library of Norway's collection}
Of the around 650\,000 books and 4.6 million newspaper issues in NLN's digitised collection, about 3000 and 4500 are classified as Sámi, respectively. The classification generally means that the texts are written in Sámi, though some may just address Sámi-related topics.

With more than 20\,000 speakers North Sámi is the most widely spoken Sámi language in Norway, Sweden and Finland, and it makes up the largest part of the Sámi collection at NLN. The other Sámi languages in NLN's collection all have less than 500 speakers. South and Lule Sámi are spoken in Norway and Sweden, and the collection contains a good amount of South and Lule Sámi books. Skolt Sámi, previously spoken in Norway and Russia, is now mainly spoken in Finland, along with Inari Sámi, which has only ever been spoken in Finland. There is much less material in these languages in the collection (\( < 20 \) books in total).

All five languages have standardised orthographies that were made or revised in the 1970s, 80s or 90s \cite{10.1093/oso/9780198767664.003.0006,olthuis2013revitalising,Magga94}, but the collection also includes earlier works that predate the standardised norms. To some extent these books contain non-standard letters or glyph-shapes and most words are spelled in ways differing from contemporary orthographies.

The Sámi written languages have letters not found in the Norwegian alphabet, but it varies from language to language which letters and how many. The alphabets have some letters in common, but none are identical. See \cref{tab:characters}
 for an overview of these characters. 

\begin{table}[h]
    \centering
    %\customfont 
    \includegraphics[]{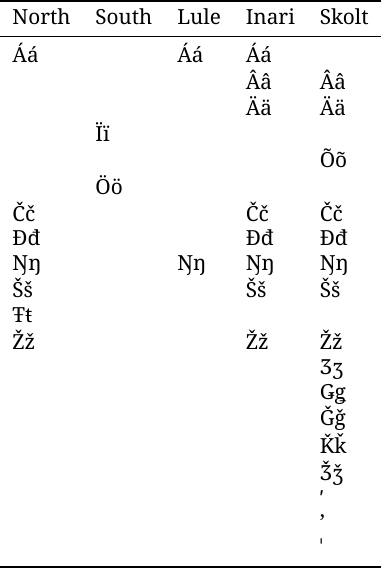}
    \caption{Overview of non-Norwegian characters used in the contemporary orthographies of the Sámi languages in the collection}
    \label{tab:characters}
\end{table}

\subsection{Related work}
While early OCR approaches often relied on hand-crafted image features combined with shape- and text-analysis \cite{smith2007overview}, modern solutions use deep learning based models to learn informative features from the data itself. In particular, developments like convolutional neural networks (CNNs), bidirectional long-short-term-memory (LSTMs) \cite{10.1162/neco.1997.9.8.1735} and the Connectionist Temporal Classification (CTC) loss \cite{graves2006connectionist} has yielded state-of-the-art results \cite{shi2016end,puigcerver2017multidimensional,vanKoert2024Loghi,10.1007/978-3-031-70549-6_23}. Recently, transformer-based machine learning advancements have led to transformer-based OCR models such as TrOCR \cite{li2023trocr}.

OCR pipelines have also been developed for collections of digitised documents: Tesseract \cite{smith2007overview} is an open-source OCR framework for line segmentation and text recognition which includes pre-trained OCR models for several languages\footnote{but none for the Sámi languages} and training scripts for training and fine-tuning on custom data. Since 2018, Tesseract has also supported LSTMs.

Another example is Transkribus, a proprietary platform for the recognition of printed and handwritten documents with a built-in interface for (semi-)manual transcription. The platform supports layout analysis and text recognition, using pre-existing or custom-trained models. The text recognition models are based on PyLaia \cite{puigcerver2017multidimensional,10.1007/978-3-031-70549-6_23}, which uses a combination of CNNs and bidirectional LSTMs. Transcriptions can be exported, though models are restricted to use within the platform.

A recent advancement is transformers-based OCR. TrOCR is a state-of-the-art text recognition model that combines powerful transformer models for vision and language \cite{li2023trocr}. Specifically, TrOCR combines the ``encoder'' of a vision transformer (ViT) \cite{dosovitskiy2021an}, with the language generating ``decoder'' of a robustly optimised Bidirectional encoder representations from transformers approach (RoBERTa) model \cite{liu2020roberta}. TrOCR is specialised for text recognition, and will not perform ancillary tasks, like layout analysis. Moreover, while TrOCR is shown capable of outperforming Transkribus and Tesseract \cite{strobel2023adaptability,li2023trocr}, it is still a relatively recent algorithm, and there is still a need to assess its accuracy for low-resource languages. 

OCR quality greatly impacts downstream processes \cite{lopresti2008optical,jarvelin2016information,evershed2014correcting}. Consequently, parts of a digitised collection with challenges like unusual fonts, bad scan quality or text in a low-resource language, will be less accessible. Several works have, thus, focused on improving OCR quality for texts with such challenges by e.g. using an ensemble of image preprocessing transforms \cite{koistinen2017improving}, comparing various OCR- or handwritten text recognition (HTR)-models for smaller languages \cite{10.1007/978-3-031-06555-2_27, memon2020handwritten, tafti2016ocr, koistinen2017improving, helinski2012report} or post-correcting outputs \cite{poncelas-etal-2020-tool,duong2020unsupervisedmethodocrpostcorrection}.

OCR for low-resource languages is particularly challenging. Not only is there much less labelled data for training, but this problem is exacerbated further by potential changes in orthographies. \newcite{rijhwani2023user} showed that including OCR in a semi-automatic annotation suite can aid annotation -- even for a low-resource language such as Kwak’wala, where automatic annotation is difficult. Similarly, \newcite{yaseen2024making} trained a Tesseract-based OCR system for Kurdish, another low-resource language. \newcite{Agarwal2024} presented a concise survey of OCR for low-resource languages with a focus on Indigenous Languages of the Americas. Finally, \newcite{partanen-riessler-2019-ocr} presented an OCR model for the Unified Northern Alphabet, used in the Soviet Union between 1931 and 1937 for Northern Minority languages (which includes Kildin Sámi).

\section{Methods}

\subsection{Data}
The main source for the data used in this work is NLN’s digitised collection. Our goal was to create an OCR model for all languages in the collection, rather than one for each language, as this would allow for the most efficient integration into NLN's digitisation pipeline. However, we realised early that including Skolt Sámi would be difficult because of the three apostrophe characters that indicate pronunciation. This makes transcription difficult without a certain level of language proficiency. Thus, we proceeded with North, South, Lule and Inari Sámi. 

In addition to data from NLN, we also used text-data data from the GiellaLT corpora\footnote{\url{https://giellalt.github.io/}} as basis for synthetic text images and data from the Divvun \& Giellatekno fork of tesstrain\footnote{\url{https://github.com/divvungiellatekno/tesstrain/tree/main/training-data/nor_sme-ground-truth}} as basis for an out-of-domain (OOD) test set. 

\subsubsection{Training data}

We trained OCR models using manually transcribed data, machine transcribed data, and synthetic data\footnote{As these texts contain copyrighted materials, the transcribed data sets can not be shared openly.}. See \cref{tab:language_distribution} for an overview.

\emphpar{Manually transcribed data}
\label{sec:manually_transcribed_data}
We used Transkribus\footnote{We used the Transkribus Expert Client v1.28.0 and \url{https://app.transkribus.org} v4.0.0.150} \cite{8270253} to create the training data from the images of scanned pages. We used the platform’s layout analysis, manually adjusting the results where necessary, then applied text recognition to the documents. Initially, we used a standard model provided by Transkribus. As we progressively corrected the recognised text, we trained new models, which were applied to recognise text in new documents, which we manually corrected to create the manually transcribed data.

Following this procedure, we transcribed 58 Sámi book and newspaper pages to create a manually transcribed training set, henceforth referred to as \emph{Ground Truth Sámi} (GT-Sámi).

\begin{table}[t]
    \centering
    \begin{threeparttable}
        \input{tables/data_distribution}
    \end{threeparttable}
    \caption{Distribution of documents and lines in each of the  Sámi languages in the different datasets. GT, Val and Test refer to the data splits of the manually annotated data. Pred is the automatically annotated dataset, Synth is the synthetic dataset (natural language text but generated images) and OOD is the OOD Giellatekno test set.}
    \label{tab:language_distribution}
\end{table}

Additionally, we already had 82 pages with 2998 manually transcribed Norwegian text lines (produced similarly as for GT-Sámi) that we included as training data. We refer to this data as \emph{Ground Truth Norwegian} (GT-Nor).

\emphpar{Synthetic data}
To add more annotated Sámi text, we created synthetic data, which we refer to as the \emph{Synthetic Sámi} dataset (Synth-Sámi). We used the SIKOR
 Sámi text corpus \cite{sikor_01.12.2021}
as a basis of well-formed Sámi text, and generated images for the text lines (adding an uppercase version for \(\simeq 10\,\%\) of the lines), using \texttt{CorpusTools}\footnote{\url{https://github.com/divvun/CorpusTools}} to parse the XML files in the \texttt{converted}-directory of the \texttt{corpus-sma}\footnote{\url{https://github.com/giellalt/corpus-sma/}}, \texttt{corpus-sme}\footnote{\url{https://github.com/giellalt/corpus-sme/}}, \texttt{corpus-smj}\footnote{\url{https://github.com/giellalt/corpus-smj/}} and \texttt{corpus-smn}\footnote{\url{https://github.com/giellalt/corpus-smn/}} repositories. The images were created with Pillow\footnote{\url{https://python-pillow.org/} (Version 10.4.0)} and Augraphy \cite{augraphy_paper}, with variation in fonts and colours, and a varying degree of imperfections and noise added, resulting in \(307\,387\) lines\footnote{Code to generate synthetic data is on GitHub: \url{https://github.com/Sprakbanken/synthetic_text_images}}.

\emphpar{Automatically transcribed data}
As mentioned earlier, we trained Transkribus models incrementally while annotating data. Eventually, our Transkribus model\footnote{\anonymize{Anonymised Transkribus model ID}{Transkribus modelID 115833, publicly available in Transkribus}} performed well on North, South and Lule Sámi, and we decided to automatically transcribe a larger amount of Sámi text with this model. We extracted page 30 from North, South and Lule Sámi books in NLN's collection and transcribed them automatically, which resulted in 2380 pages forming the \emph{Predicted Sámi} (Pred-Sámi) dataset. This boosted the amount of data, but naturally, the transcriptions may not be correct.

\subsubsection{Validation data}
To evaluate during training and to select the best performing models for each architecture, we created a validation dataset. This dataset consists of 25 pages manually transcribed following the procedure described for GT-Sámi. Lines were selected from different books than the GT-Sámi training data while keeping a similar language distribution.
\subsubsection{Test data}
To compare the OCR approaches we used two test sets: one from NLN's collection and one from Divvun \& Giellatekno’s tesstrain data.
\emphpar{NLN test data}
As a goal of this work was to improve the transcriptions of Sámi documents in NLN's collection, we created a test set based on current transcriptions (baseline) of 21 pages from 18 books and 2 newspapers provided by NLN\footnote{We chose distinct books for the train, validation and test sets. However, due to few Inari Sámi books, 1 book is in both the train and test sets and 2 are in both the validation and test sets, but there is no page-overlap.}. NLN stores these transcriptions as Analyzed Layout and Text Object-Extensible Markup Language (ALTO-XML) files with line segmentations and transcriptions. By matching the ALTO-XML transcriptions with manually annotated data, we created a test-set containing 848 text-lines.
\emphpar{Giellatekno test data}
The Giellatekno test data \textit{nor-sme} was made for evaluating OCR reading of dictionares. It consists of 122 lines of dictionary data, thus text both in Norwegian and (contemporary) North Sámi. The dataset is available on Giellatekno's GitHub\footnote{\url{https://github.com/divvungiellatekno/tesstrain/tree/main/training-data/nor_sme-ground-truth}. We have corrected four transcriptions and used our corrected version of the test set which can be found on \url{https://github.com/MarieRoald/tesstrain/tree/fix-transcriptions/training-data/nor_sme-ground-truth} } We refer to this dataset as the OOD Giellatekno test set.
%\todo[inline]{
%Testsett basert på giellatekno \href{https://github.com/divvungiellatekno/tesstrain/tree/main/training-data}{data}

%\togiellatekno[inline]{nor-sme, forthcoming..}

\subsection{Evaluation metrics} \label{sec:eval.metrics}
Following previous work \cite{neudecker2021survey,Agarwal2024}, we used the character error rate (CER) and word error rate (WER) evaluation metrics. Specifically, we calculated collection level CER and WER (concatenating lines, with a space to separate them for WER) with Jiwer\footnote{\url{https://github.com/jitsi/jiwer} (Version 3.0.4)}.

We also calculated an \(\text{F}_1\) score for characters specific to the different Sámi languages, and an overall \(\text{F}_1\) score for all non-Norwegian Sámi characters. The \(\text{F}_1\)  score is given by  \(\text{F}_1  =2 \text{TP}/(2 \text{TP} + \text{FN} + \text{FP})\), where {TP}, {FP} and {FN} is the number of true positives, false positives and false negatives, respectively. To measure {TP}, {FP} and {FN} in an OCR-setting, we only considered character counts, not location. Thus, for a given character, \(c\), we set \(\text{TP}_c = \min(n_c^{(\text{true})}, n_c^{(\text{pred})})\), \(\text{FN} = \max(n_c^{(\text{true})} - n_c^{(\text{pred})}, 0)\) and \(\text{FP} = \max(n_c^{(\text{pred})} -  n_c^{(\text{true})}, 0)\), where \(n_c^{(\text{true})}\) and \(n_c^{(\text{pred})}\) are the number of \(c\)  characters in the ground truth and predicted transcriptions, respectively. To compute an overall \(\text{F}_1\), we combined the {TP}, {FN}, and {FP} across all lines and characters-of-interest.

To examine the types of errors our models made, we calculated the most common errors. Specifically, we used Stringalign \cite{Moe_Stringalign_2024}, which implements optimal string alignment. Note that, in theory, multiple alignments can exist (e.g. if two letters are swapped), in which case Stringalign picks one.

\subsection{Models and training}
\label{sec:Models}
A goal of this work was evaluating different state-of-the-art OCR frameworks for Sámi text recognition. Specifically, we compared Transkribus, Tesseract and TrOCR. For each approach, we trained on several dataset combinations and chose the model based on mean(CER, WER) on the validation data for test-set evaluation.

\subsubsection{Transkribus}
We used Transkribus Expert for training Transkribus models\footnote{\url{https://help.transkribus.org/model-setup-and-training}}. We used standard parameters, but opted ``Using exsisting line polygons for training'', and changed the batch size from 24 to 12\footnote{We changed this parameter after advice from the Transkribus team due to problems with the training stopping with \texttt{exitCode = 1}}. We set 100 as maximum numbers of epochs, and 20 as early stopping. We used Transkribus print M1\footnote{Transkribus ModelID 39995} as base model for 4 of the 5 models.  All Transkribus models were run with the setting ``Use language model''\footnote{Which uses PyLaia's n-gram model functionality to inform character predictions \cite{10.1007/978-3-031-70549-6_23}.}.

\subsubsection{Tesseract}
We used the official tesstrain repository\footnote{\url{https://github.com/tesseract-ocr/tesstrain} (Version 1.0.0, commit hash \href{https://github.com/tesseract-ocr/tesstrain/commit/45cacc5c05929a0d8a19d19104d4a0718877a91c}{45cacc5})} and Tesseract 5.4.1 for training. We experimented with both training models from scratch and fine-tuning existing models. During early experiments, we tried fine-tuning Norwegian, Finnish, and Estonian models using our Sámi dataset, and observed that the model with the Norwegian base adapted faster and performed better on our validation set. Thus, we continued training with the Norwegian base\footnote{\url{https://github.com/tesseract-ocr/tessdata_best/blob/main/nor.traineddata}}.

As tesstrain does not support dynamic learning rate and only exposes a few training hyperparameters to the user, we trained our models in 1-20 epoch increments, updating the learning rate until the model checkpoints no longer showed improvements on the validation set.

\subsubsection{TrOCR}
We used Huggingface Transformers \cite{wolfTransformersStateoftheArtNatural2020} to fit the TrOCR models, initialising with the parameters from the \texttt{microsoft/trocr-base-printed} repository. This model is pre-trained on both synthetic and printed text \cite{li2023trocr}. For fine-tuning, we had an initial learning rate of \(10^{-6}\), decreasing it by a constant amount for each iteration until it reached \(10^{-7}\) at the final iteration. For models fine-tuned without Pred-Sámi, we trained for 200 epochs, evaluating and storing model parameters every fifth epoch. However, due to the data size and hardware limitations, models that included Pred-Sámi were only fine-tuned for 100 epochs, evaluating and storing model parameters every second epoch and selecting the checkpoint with the lowest validation CER.

\begin{table*}
    \centering
    \input{tables/valset_cer_wer2}
    \caption{CER, WER, and mean(CER, WER) on the validation set. The checkmarks indicate whether models were trained from scratch (i.e. not fine-tuning an existing base model) (first column) and what datasets were part of the training data}
    \label{tab:val_set_cer_wer}
\end{table*}

\subsubsection{Pre-training with synthetic data}
We trained additional TrOCR and Tesseract models using synthetic data to assess the effect of adding such data\footnote{We did not train Transkribus models with synthetic data as it does not support an easy way to train based on line images and because of its page-based pricing model.}. After training all models without synthetic data, we retrained with the smallest amount of hand-annotated data (GT-Sámi) and best performing data combination, this time initialising with a model pre-trained on Synth-Sámi. 

In particular, due to time and hardware limitations, we trained models on synthetic data in two stages inspired by the two-stage procedure in e.g \cite{li2023trocr}. For the first stage, we trained for five epochs on Synth-Sámi. For the second stage, we initialised with the best checkpoint from the first stage (lowest CER) and continued training on real data.

\section{Results}

Code for training Tesseract and TrOCR models, creating synthetic data and more detailed dataset information is available through the supplement on GitHub\footnote{\url{\anonymize{https://github.com/anonymised/repo-name}{https://github.com/Sprakbanken/nodalida25_sami_ocr}}}.
\subsection{NLN validation data}

\subsubsection{Transkribus models}
As shown in \cref{tab:val_set_cer_wer}, CER and WER decreased when we used the Transkribus Print M1 as the base model in addition to GT-Sámi. Hence, we continued to use the base model in the subsequent training. Supplementing GT-Sámi with GT-Nor did not improve performance, while supplementing with Pred-Sámi increased CER but decreased WER. However, adding both GT-Nor and Pred-Sámi led to the best-performing model on the validation set. 

\subsubsection{Tesseract models}
From \cref {tab:val_set_cer_wer}, we see that the model trained on GT-Sámi with a Norwegian base model greatly outperformed the corresponding model without a base model. We therefore continued training all Tesseract models from the Norwegian base model. Adding GT-Nor to the training data worsened the validation performance. However, adding Pred-Sámi to the training data improved validation performance, and adding both further improved the performance. Using Synth-Sámi also improved performance, and the model performed best in terms of mean(CER, WER) when all training datasets were used.

\subsubsection{TrOCR models}
For TrOCR, we observed that including GT-Nor in the training had a slight improvement when only training with GT-Sámi and no improvement when training with GT-Sámi and Pred-Sámi (see \cref{tab:val_set_cer_wer}). Moreover, while including Pred-Sámi improved performance, pre-training with Synth-Sámi had a larger effect. The overall best-performing model was trained with both Synth-Sámi and Pred-Sámi in addition to GT-Sámi.

\begin{table*}[t]
    \centering
    \input{tables/test_performance}
    \caption{CER, WER and Sámi letter F1 on NLN test data. The score for each language and overall score across languages are listed. Transkribus, Tesseract and TrOCR refer to the best performing model on the validation set for each model type. Baseline is the current OCR output in the online library. The downward arrows indicate that a low score is better, while the upward arrow indicates that a high score is better.}
    \label{tab:test_set_performance}
\end{table*}

\begin{table*}[t]
    \centering
    \input{tables/character_errors}
    \vspace*{1ex}\newline
    {
    \begin{tabular}{c@{}c@{}c@{: }l@{\hspace{1em}}r@{: }l}
     `a' & \(\shortrightarrow\) & `b' & model transcribed ``a'' as ``b'' & \(n_e\) & Error count\\
     `a' & \(\shortrightarrow\) & `' & model incorrectly deleted ``a'' & \(n_m\) & Misses of the character left of \(\shortrightarrow\)\\
     `' & \(\shortrightarrow\) & `b' & model incorrectly inserted ``b'' & \(n_c\) & Occurrences of the character left of \(\shortrightarrow\)
    \end{tabular}
    }
    %\begin{tabular}{c@{}c@{}c@{: }l}
    % `a' & \(\rightarrow\) & `b' & model transcribed ``a'' as ``b'' \\
     %`a' & \(\rightarrow\) & `' & model incorrectly deleted ``a'' \\
    % `' & \(\rightarrow\) & `b' & model incorrectly inserted ``b''
    %\end{tabular}
    \caption{Top ten most common errors on the NLN test data. Transkribus, Tesseract and TrOCR refers to the best performing model on the validation set for each model type. Baseline is the current OCR output in the online library.}
    \label{tab:test.character.errors}
\end{table*}

\subsection{NLN test data}
\Cref{tab:test_set_performance}, shows that while Transkribus achieves a lower CER for most languages, it obtains a higher WER and a lower special character \(\text{F}_1\)-score compared to TrOCR. Tesseract performed worst on this dataset. However, all models greatly improve compared to the baseline, with the CER and WER being reduced by factors between 3.8 and 5.6.

The special character \(\text{F}_1\)-score in \cref{tab:test_set_performance} shows that the baseline struggles with non-Norwegian Sámi characters. While the \(\text{F}_1\) score does not take letter position into account, we also see the same pattern reflected in \cref{tab:test.character.errors}, which shows that seven of the ten most common mistakes for the baseline are replacing a non-Norwegian Sámi special character. In contrast, we see that our three models make fewer mistakes, and their ten most common mistakes are less systematically replacing distinctive Sámi characters and include, e.g. insertions and deletions.

\subsection{Giellatekno test data}
In contrast to the NLN test data, the Tesseract model performed the best on the OOD test data from Giellatekno for all metrics (see \cref{tab:giellatek.testset}). Transkribus was worst in terms of CER and WER, while TrOCR was worst in terms of the \(\text{F}_1\) score.

In \cref{tab:gieallatekno.character.errors}, we see the most common errors on the Giellatekno test set. The Transkribus model seems to have a tendency to add punctuation marks, and mistake the letter ø for e. All models fail to transcribe ü (of which there are only two in the Giellatekno test set). This is not surprising, as the letter rarely appears in the training data \footnote{The letter ü appears 59 times in Synth-Sámi, 9 times in Pred-Sámi and 5 times in GT-Nor.}.

\begin{table}[ht]
    \centering
    \input{tables/giellatekno_testset_performance}

    \caption{CER, WER and Sámi letter \(\text{F}_1\) on the OOD Giellatekno test set. The downwards arrows indicate that a low score is better, while the upwards arrow indicates that a high score is better.}
    \label{tab:giellatek.testset}
\end{table}

\begin{table*}[t]
    \centering
    \begin{tabular}{c}
    \input{tables/giellatekno_character_errors}\vspace*{1ex}\\
    {
    \begin{tabular}{c@{}c@{}c@{: }l@{\hspace{1em}}r@{: }l}
     `a' & \(\shortrightarrow\) & `b' & model transcribed ``a'' as ``b'' & \(n_e\) & Error count\\
     `a' & \(\shortrightarrow\) & `' & model incorrectly deleted ``a'' & \(n_m\) & Misses of the character left of \(\shortrightarrow\)\\
     `' & \(\shortrightarrow\) & `b' & model incorrectly inserted ``b'' & \(n_c\) & Occurrences of the character left of \(\shortrightarrow\)
    \end{tabular}
    }
    \end{tabular}
    \caption{Top ten most common errors on the OOD Giellatekno test data. Transkribus, Tesseract and TrOCR refers to the best performing model on the validation set for each model type.}
    \label{tab:gieallatekno.character.errors}
\end{table*}

\section{Discussion and conclusions}
From \cref{tab:val_set_cer_wer,tab:test_set_performance}, we observe a jump in performance for the test set compared to the validation set. This increase is expected, as the test set annotations are of higher quality (more accurate line segmentations).

We see that applying a two-stage training using synthetic data for the first stage always improved the results. As such, if manual annotations are limited, the addition of synthetic data is worth considering. Furthermore, while the Pred-Sámi improved performance, its effect was less than including synthetic data. It would, thus, be interesting to investigate if further training on Synth-Sámi could eliminate the effect of Pred-Sámi. Finally, we note that including GT-Nor had a minimal effect when combined with Pred-Sámi. This finding, combined with the effect of pre-trained base models, suggests that language-independent features are already learned by the base models and highlights the value of language-specific data for fine-tuning on low-resource languages.

Unfortunately, as this work focuses on low-resource languages, few digitised texts exist. There is, therefore, a slight overlap between the books (but not pages) in the test set and the validation and training sets for Inari Sámi which could bias our results for the Inari Sámi language. Still, Inari Sámi obtained the worst CER and WER for Transkribus and the worst CER and second worst WER for Tesseract. Despite low amount of Inari Sámi, we included it in our analysis as there is an overlap between this alphabet and the North Sámi alphabet, and our OCR models could improve upon NLN's transcription for Inari Sámi.

All models improved considerably compared to the baseline and are good candidates for a re-OCR process. If transcription accuracy is the main focus, then Transkribus appears to perform the best. However, while Tesseract achieved the worst performance for the NLN test set, it performed the best on the OOD Giellatekno test set. Tesseract also has other benefits: it is available as open-source software and requires less compute than a TrOCR model.

While language-specific annotations are valuable, they are demanding to create, particularly for low-resource languages without good base models for semi-automatic annotations. However, our results show that by fine-tuning pre-trained models and augmenting manually annotated data with machine-annotated data and synthetic text images, we can achieve accurate OCR for Sámi languages, even with modest amounts of manual annotations.

\section{Further work}
As NLN's collection includes works predating the standardised Sámi orthographies, a more accurate evaluation of the OCR could be gained by examining performance across different time periods. Moreover, training specialised models to transcribe non-standard letters or glyph-shapes could enable more detailed down-stream studies of changes in orthographies. Another gap is training OCR for other Sámi languages, such as Skolt Sámi.

Given that our results show that initialising on a dataset of synthetic text images was beneficial, it is worth exploring further. The models in this work are only trained on synthetic data for five epochs, indicating that potential improvements could be made by training on synthetic data for longer, i.e. until convergence. Moreover, creating a larger synthetic dataset with greater variation of text, fonts and augmentations (e.g. additional scanning augmentations or simulating non-standard orthographies), could improve the results further. 

As this study focuses on the text recognition step of the OCR pipeline and compares three models, future research should explore additional OCR components and models. E.g. examining the effect of different line segmentation models and assessing if performance can be improved by fine-tuning the line segmentation or using end-to-end models. Additionally, extending the range of models examined --- to include tools such as PyLaia \cite{puigcerver2017multidimensional,10.1007/978-3-031-70549-6_23} (which is part of Transkribus’ pipeline), Loghi \cite{vanKoert2024Loghi}, GOT-OCR \cite{wei2024generalocrtheoryocr20} or larger TrOCR models --- could yield improvements. Lastly, including post processing, e.g. with tools from GiellaLT \cite{pirinen2023giellalt}, could improve OCR quality.

\section*{Acknowledgments}
We would like to thank Arne Martinus Lindstad for his contributions to the annotation of the data and valuable feedback.

\bibliographystyle{acl_natbib}
\bibliography{SamiskOCR}

\end{document}

%% file: tables/data_distribution.tex
\begin{tabular}{@{}l@{\hspace{0.5em}}lrrrr@{}}
\toprule
 &  & South & North & Lule & Inari \\
\midrule
\multirow[c]{4}{*}{\vspace*{-2ex}\rotatebox{90}{Docs}} & GT & 5 & 3 & 2 & 3 \\
 & Pred & 265 & 1810 & 235 & 0 \\
 & Val & 2 & 8 & 2 & 3 \\
 & Test & 4 & 7 & 4 & 5 \\
\addlinespace[0.5ex] % Adds 0.5cm of space
\cline{1-6}
\addlinespace[1ex] % Adds 0.5cm of space after the cline
\multirow[c]{6}{*}{\vspace*{-2ex}\rotatebox{90}{Lines}} & GT & 208 & 5572 & 81 & 280 \\
 & Pred & 7082 & 70413 & 6781 & 0 \\
 & Synth & 76971 & 76949 & 76970 & 76497 \\
 & Val & 53 & 1837 & 36 & 109 \\
 & Test & 195 & 353 & 137 & 163 \\
 & OOD & 0 & 122 & 0 & 0 \\
\bottomrule
\end{tabular}

%% file: tables/valset_cer_wer2.tex
\begin{tabular}{lllllccccccccc}
\toprule
\multirow{4}*{\rotatebox{90}{w/o base}} & \multirow{4}*{\rotatebox{90}{GT-Sámi}} & \multirow{4}*{\rotatebox{90}{GT-Nor}} & \multirow{4}*{\rotatebox{90}{Pred-Sámi}} & \multirow{4}*{\rotatebox{90}{Synth base}}& \\
&&&&&&&&&&&&& \\ 
&&&&& \multicolumn{3}{c}{\textbf{Transkribus}} & \multicolumn{3}{c}{\textbf{Tesseract}} & \multicolumn{3}{c}{\textbf{TrOCR}} \\
&&&& & CER & WER & mean & CER & WER & mean & CER & WER & mean \\
\cmidrule(r){1-5}\cmidrule(lr){6-8}\cmidrule(lr){9-11}\cmidrule(l){12-14}
\checkmark & \checkmark &  &  &  & 1.59 & 5.67 & 3.63 & 5.53 & 24.70 & 15.11 &  &  &  \\
\rowcolor{gray!20}  & \checkmark &  &  &  & 1.28 & 4.34 & 2.81 & 2.05 & 9.84 & 5.95 & 1.98 & 9.29 & 5.64 \\
 & \checkmark & \checkmark &  &  & 1.31 & 4.35 & 2.83 & 2.37 & 11.39 & 6.88 & 1.95 & 8.88 & 5.42 \\
\rowcolor{gray!20} & \checkmark &  & \checkmark &  & 1.48 & 4.02 & 2.75 & 1.85 & 8.17 & 5.01 & 1.28 & 5.00 & 3.14 \\
& \checkmark & \checkmark & \checkmark &  & \textbf{1.07} & \textbf{3.58} & \textbf{2.33} & 1.81 & 7.96 & 4.89 & 1.32 & 5.14 & 3.23 \\
\rowcolor{gray!20} & \checkmark &  &  & \checkmark &  &  &  & \textbf{1.78} & 8.78 & 5.28 & 1.15 & 5.04 & 3.09 \\
& \checkmark &  & \checkmark & \checkmark &  &  &  &  &  &  & \textbf{1.08} & \textbf{4.29} & \textbf{2.69} \\
\rowcolor{gray!20} & \checkmark & \checkmark & \checkmark & \checkmark &  &  &  & 1.79 & \textbf{7.70} & \textbf{4.75} &  &  &  \\
\bottomrule
\end{tabular}

%% file: tables/test_performance.tex
\begin{tabular}{llrrrr}
\toprule
 &  & Transkribus & Tesseract & TrOCR & Baseline \\
\midrule
\multirow[t]{5}{*}{CER \(\downarrow\) [\(\%\)]} & Overall & {\cellcolor[HTML]{2F974E}} \color[HTML]{F1F1F1} 0.61 & {\cellcolor[HTML]{81CA81}} \color[HTML]{000000} 0.89 & {\cellcolor[HTML]{50B264}} \color[HTML]{F1F1F1} 0.74 & 3.38 \\
 & South & {\cellcolor[HTML]{005723}} \color[HTML]{F1F1F1} 0.33 & {\cellcolor[HTML]{BBE4B4}} \color[HTML]{000000} 1.09 & {\cellcolor[HTML]{005723}} \color[HTML]{F1F1F1} 0.33 & 2.05 \\
 & North & {\cellcolor[HTML]{1E8741}} \color[HTML]{F1F1F1} 0.53 & {\cellcolor[HTML]{4BB062}} \color[HTML]{F1F1F1} 0.73 & {\cellcolor[HTML]{D4EECE}} \color[HTML]{000000} 1.20 & 3.99 \\
 & Lule & {\cellcolor[HTML]{005A24}} \color[HTML]{F1F1F1} 0.34 & {\cellcolor[HTML]{00441B}} \color[HTML]{F1F1F1} 0.26 & {\cellcolor[HTML]{39A257}} \color[HTML]{F1F1F1} 0.66 & 2.46 \\
 & Inari & {\cellcolor[HTML]{D9F0D3}} \color[HTML]{000000} 1.22 & {\cellcolor[HTML]{F7FCF5}} \color[HTML]{000000} 1.43 & {\cellcolor[HTML]{067230}} \color[HTML]{F1F1F1} 0.43 & 4.36 \\
\midrule
\multirow[t]{5}{*}{WER \(\downarrow\) [\(\%\)]} & Overall & {\cellcolor[HTML]{268E47}} \color[HTML]{F1F1F1} 3.19 & {\cellcolor[HTML]{7AC77B}} \color[HTML]{000000} 4.65 & {\cellcolor[HTML]{1C8540}} \color[HTML]{F1F1F1} 2.96 & 18.71 \\
 & South & {\cellcolor[HTML]{016E2D}} \color[HTML]{F1F1F1} 2.42 & {\cellcolor[HTML]{F7FCF5}} \color[HTML]{000000} 7.45 & {\cellcolor[HTML]{00692A}} \color[HTML]{F1F1F1} 2.33 & 15.98 \\
 & North & {\cellcolor[HTML]{00441B}} \color[HTML]{F1F1F1} 1.66 & {\cellcolor[HTML]{18823D}} \color[HTML]{F1F1F1} 2.90 & {\cellcolor[HTML]{2F984F}} \color[HTML]{F1F1F1} 3.41 & 20.08 \\
 & Lule & {\cellcolor[HTML]{2A924A}} \color[HTML]{F1F1F1} 3.27 & {\cellcolor[HTML]{004D1F}} \color[HTML]{F1F1F1} 1.84 & {\cellcolor[HTML]{329B51}} \color[HTML]{F1F1F1} 3.47 & 13.27 \\
 & Inari & {\cellcolor[HTML]{CEECC8}} \color[HTML]{000000} 6.18 & {\cellcolor[HTML]{EFF9EC}} \color[HTML]{000000} 7.13 & {\cellcolor[HTML]{006D2C}} \color[HTML]{F1F1F1} 2.40 & 22.62 \\
\midrule
\multirow[t]{5}{*}{Sámi letter \(\text{F}_1\) \(\uparrow\) [\(\%\)]} & Overall & {\cellcolor[HTML]{0E7936}} \color[HTML]{F1F1F1} 96.03 & {\cellcolor[HTML]{339C52}} \color[HTML]{F1F1F1} 93.81 & {\cellcolor[HTML]{006B2B}} \color[HTML]{F1F1F1} 96.97 & 52.54 \\
 & South & {\cellcolor[HTML]{84CC83}} \color[HTML]{000000} 90.24 & {\cellcolor[HTML]{F7FCF5}} \color[HTML]{000000} 83.02 & {\cellcolor[HTML]{319A50}} \color[HTML]{F1F1F1} 93.92 & 24.52 \\
 & North & {\cellcolor[HTML]{00491D}} \color[HTML]{F1F1F1} 98.57 & {\cellcolor[HTML]{006729}} \color[HTML]{F1F1F1} 97.13 & {\cellcolor[HTML]{006428}} \color[HTML]{F1F1F1} 97.27 & 55.85 \\
 & Lule & {\cellcolor[HTML]{005723}} \color[HTML]{F1F1F1} 97.91 & {\cellcolor[HTML]{005723}} \color[HTML]{F1F1F1} 97.88 & {\cellcolor[HTML]{00682A}} \color[HTML]{F1F1F1} 97.06 & 51.75 \\
 & Inari & {\cellcolor[HTML]{258D47}} \color[HTML]{F1F1F1} 94.70 & {\cellcolor[HTML]{3CA559}} \color[HTML]{F1F1F1} 93.22 & {\cellcolor[HTML]{00441B}} \color[HTML]{F1F1F1} 98.84 & 68.61 \\
%\cline{1-6}
\bottomrule
\end{tabular}

%% file: tables/character_errors.tex
\setlength{\tabcolsep}{3.7pt}
\setlength{\cmidrulekern}{3.7pt}
\begin{tabular}{@{}c@{}c@{}crrr|c@{}c@{}crrr|c@{}c@{}crrr|c@{}c@{}crrr@{}}
\toprule
\multicolumn{6}{c}{\textbf{Transkribus}} & \multicolumn{6}{c}{\textbf{Tesseract}} & \multicolumn{6}{c}{\textbf{TrOCR}} & \multicolumn{6}{c}{\textbf{Baseline}} \\
\cmidrule(r){1-6}\cmidrule(lr){7-12}\cmidrule(lr){13-18}\cmidrule(l){19-24}
\multicolumn{3}{c}{Error} & \(n_e\) & \(n_m\) & \multicolumn{1}{c}{\(n_c\)} & \multicolumn{3}{c}{Error} & \(n_e\) & \(n_m\) & \multicolumn{1}{c}{\(n_c\)} & \multicolumn{3}{c}{Error} & \(n_e\) & \(n_m\) & \multicolumn{1}{c}{\(n_c\)} & \multicolumn{3}{c}{Error} & \(n_e\) & \(n_m\) & \multicolumn{1}{c}{\(n_c\)} \\
\cmidrule(r){1-3}\cmidrule(lr){4-4}\cmidrule(lr){5-5}\cmidrule(lr){6-6}
\cmidrule(lr){7-9}\cmidrule(lr){10-10}\cmidrule(lr){11-11}\cmidrule(lr){12-12}
\cmidrule(lr){13-15}\cmidrule(lr){16-16}\cmidrule(lr){17-17}\cmidrule(lr){18-18}
\cmidrule(lr){19-21}\cmidrule(lr){22-22}\cmidrule(lr){23-23}\cmidrule(l){24-24}
`â' & \(\shortrightarrow\) & `á' & 16 & 35 & 287 & `ï' & \(\shortrightarrow\) & `i' & 24 & 27 & 160 & `Á' & \(\shortrightarrow\) & `A' & 9 & 11 & 28 & `á' & \(\shortrightarrow\) & `å' & 313 & 418 & 1136 \\
`â' & \(\shortrightarrow\) & `a' & 14 & 35 & 287 & `â' & \(\shortrightarrow\) & `á' & 22 & 29 & 287 & `' & \(\shortrightarrow\) & `l' & 7 & -- & -- & `ï' & \(\shortrightarrow\) & `i' & 137 & 139 & 160 \\
`Á' & \(\shortrightarrow\) & `A' & 9 & 10 & 28 & `đ' & \(\shortrightarrow\) & `d' & 12 & 14 & 173 & `Š' & \(\shortrightarrow\) & `S' & 6 & 6 & 6 & `â' & \(\shortrightarrow\) & `å' & 103 & 180 & 287 \\
`/' & \(\shortrightarrow\) & ` ' & 9 & 9 & 10 & `Á' & \(\shortrightarrow\) & `A' & 10 & 11 & 28 & `' & \(\shortrightarrow\) & `i' & 5 & -- & -- & `–' & \(\shortrightarrow\) & `-' & 75 & 77 & 82 \\
`i' & \(\shortrightarrow\) & `ï' & 7 & 13 & 3299 & `' & \(\shortrightarrow\) & `d' & 8 & -- & -- & `' & \(\shortrightarrow\) & ` ' & 4 & -- & -- & `š' & \(\shortrightarrow\) & `s' & 72 & 95 & 215 \\
`đ' & \(\shortrightarrow\) & `d' & 7 & 11 & 173 & `' & \(\shortrightarrow\) & `á' & 7 & -- & -- & `i' & \(\shortrightarrow\) & `ï' & 4 & 21 & 3299 & `đ' & \(\shortrightarrow\) & `d' & 48 & 61 & 173 \\
`š' & \(\shortrightarrow\) & `' & 6 & 6 & 215 & `' & \(\shortrightarrow\) & `i' & 7 & -- & -- & `á' & \(\shortrightarrow\) & `å' & 4 & 14 & 1136 & `á' & \(\shortrightarrow\) & `a' & 46 & 418 & 1136 \\
`ä' & \(\shortrightarrow\) & `á' & 5 & 6 & 150 & `s' & \(\shortrightarrow\) & `S' & 7 & 8 & 1509 & `Č' & \(\shortrightarrow\) & `C' & 4 & 4 & 8 & `â' & \(\shortrightarrow\) & `á' & 30 & 180 & 287 \\
`ï' & \(\shortrightarrow\) & `i' & 5 & 5 & 160 & `â' & \(\shortrightarrow\) & `å' & 6 & 29 & 287 & `á' & \(\shortrightarrow\) & `a' & 3 & 14 & 1136 & `â' & \(\shortrightarrow\) & `ä' & 26 & 180 & 287 \\
`' & \(\shortrightarrow\) & `-' & 4 & -- & -- & `.' & \(\shortrightarrow\) & `' & 5 & 6 & 509 & `a' & \(\shortrightarrow\) & `u' & 3 & 8 & 3247 & `č' & \(\shortrightarrow\) & `c' & 26 & 62 & 163 \\
\bottomrule
\end{tabular}

%% file: tables/giellatekno_testset_performance.tex
\setlength{\tabcolsep}{0.3em}
\begin{tabular}{lrrr}
\toprule
 & Transkribus & Tesseract & TrOCR \\
\midrule
CER \(\downarrow\) [\(\%\)] & {\cellcolor[HTML]{F7FCF5}} \color[HTML]{000000} 0.70 & {\cellcolor[HTML]{00441B}} \color[HTML]{F1F1F1} 0.12 & {\cellcolor[HTML]{80CA80}} \color[HTML]{000000} 0.43 \\
WER \(\downarrow\) [\(\%\)] & {\cellcolor[HTML]{F7FCF5}} \color[HTML]{000000} 5.85 & {\cellcolor[HTML]{00441B}} \color[HTML]{F1F1F1} 1.02 & {\cellcolor[HTML]{6ABF71}} \color[HTML]{000000} 3.31 \\
F1 \(\uparrow\) [\(\%\)] & {\cellcolor[HTML]{00441B}} \color[HTML]{F1F1F1} 100.00 & {\cellcolor[HTML]{00441B}} \color[HTML]{F1F1F1} 100.00 & {\cellcolor[HTML]{F7FCF5}} \color[HTML]{000000} 98.33 \\
\bottomrule
\end{tabular}

%% file: tables/giellatekno_character_errors.tex
\setlength{\tabcolsep}{3.7pt}
\setlength{\cmidrulekern}{3.7pt}
\begin{tabular}{@{}c@{}c@{}crrr|c@{}c@{}crrr|c@{}c@{}crrr@{}}
\toprule
\multicolumn{6}{c}{\textbf{Transkribus}} & \multicolumn{6}{c}{\textbf{Tesseract}} & \multicolumn{6}{c}{\textbf{TrOCR}} \\
\cmidrule(r){1-6}\cmidrule(lr){7-12}\cmidrule(l){13-18}
\multicolumn{3}{c}{Error} & \(n_e\) & \(n_m\) & \multicolumn{1}{c}{\(n_c\)} & \multicolumn{3}{c}{Error} & \(n_e\) & \(n_m\) & \multicolumn{1}{c}{\(n_c\)} & \multicolumn{3}{c}{Error} & \(n_e\) & \(n_m\) & \multicolumn{1}{c}{\(n_c\)} \\
\cmidrule(r){1-3}\cmidrule(lr){4-4}\cmidrule(lr){5-5}\cmidrule(lr){6-6}
\cmidrule(lr){7-9}\cmidrule(lr){10-10}\cmidrule(lr){11-11}\cmidrule(lr){12-12}
\cmidrule(lr){13-15}\cmidrule(lr){16-16}\cmidrule(lr){17-17}\cmidrule(l){18-18}
`' & \(\shortrightarrow\) & `.' & 12 & -- & -- & `ü' & \(\shortrightarrow\) & `i' & 1 & 2 & 2 & `ü' & \(\shortrightarrow\) & `ï' & 2 & 2 & 2 \\
`ø' & \(\shortrightarrow\) & `e' & 4 & 5 & 13 & `ü' & \(\shortrightarrow\) & `u' & 1 & 2 & 2 & `' & \(\shortrightarrow\) & `,' & 1 & -- & -- \\
`' & \(\shortrightarrow\) & `,' & 2 & -- & -- & `t' & \(\shortrightarrow\) & `f' & 1 & 1 & 220 & `t' & \(\shortrightarrow\) & `l' & 1 & 2 & 220 \\
`ü' & \(\shortrightarrow\) & `u' & 2 & 2 & 2 & `n' & \(\shortrightarrow\) & `m' & 1 & 1 & 164 & `te' & \(\shortrightarrow\) & `s' & 1 & 2 & 28 \\
`' & \(\shortrightarrow\) & `k' & 1 & -- & -- &  &  &  &  &  &  & `l' & \(\shortrightarrow\) & `' & 1 & 1 & 169 \\
`ø' & \(\shortrightarrow\) & `o' & 1 & 5 & 13 &  &  &  &  &  &  & `o' & \(\shortrightarrow\) & `n' & 1 & 1 & 149 \\
`c' & \(\shortrightarrow\) & `' & 1 & 1 & 23 &  &  &  &  &  &  & `m' & \(\shortrightarrow\) & `n' & 1 & 1 & 69 \\
 &  &  &  &  &  &  &  &  &  &  &  & `c' & \(\shortrightarrow\) & `e' & 1 & 1 & 23 \\
 &  &  &  &  &  &  &  &  &  &  &  & `-' & \(\shortrightarrow\) & `–' & 1 & 1 & 18 \\
 &  &  &  &  &  &  &  &  &  &  &  & `\textipa{N}' & \(\shortrightarrow\) & `ž' & 1 & 1 & 9 \\
 &  &  &  &  &  &  &  &  &  &  &  & `=' & \(\shortrightarrow\) & `2' & 1 & 1 & 4 \\
 &  &  &  &  &  &  &  &  &  &  &  & `x' & \(\shortrightarrow\) & `s' & 1 & 1 & 2 \\
\bottomrule
\end{tabular}